\documentclass[runningheads]{llncs}
\usepackage{graphicx}
\usepackage{multirow} 
\usepackage{xcolor}
\usepackage{float} 
\usepackage{subfigure}
\usepackage[ruled,linesnumbered]{algorithm2e}
\begin{document}
\title{Graph Federated Learning Based on the Decentralized Framework}
\author{Peilin Liu\inst{1} \and
Yanni Tang\inst{2, 3} \and
Mingyue Zhang\inst{3} \and
Wu Chen\inst{3}}
\authorrunning{Peilin. L et al.}
\institute{College of Computer and Information Science \& College of Software, Southwest University, Chongqing, 400715, China \and
School of Computer Science, University of Auckland, Auckland, 1142, New Zealand \and
College of Software, Southwest University, Chongqing, 400715, China
}
\maketitle              % typeset the header of the contribution
\begin{abstract}

Graph learning has a wide range of applications in many scenarios, which require more need for data privacy. Federated learning is an emerging distributed machine learning approach that leverages data from individual devices or data centers to improve the accuracy and generalization of the model, while also protecting the privacy of user data. Graph-federated learning is mainly based on the classical federated learning framework i.e., the Client-Server framework. However, the Client-Server framework faces problems such as a single point of failure of the central server and poor scalability of network topology. First, we introduce the decentralized framework to graph-federated learning. Second, determine the confidence among nodes based on the similarity of data among nodes, subsequently, the gradient information is then aggregated by linear weighting based on confidence. Finally, the proposed method is compared with FedAvg, Fedprox, GCFL, and GCFL+ to verify the effectiveness of the proposed method. Experiments demonstrate that the proposed method outperforms other methods.

\end{abstract}

\section{Introduction}
% version 1；
% 0. 简述图学习，在医疗保健、金融、智能交通和工业生产等领域都有应用->在这些场景下数据隐私增->FL
%  1. 数据隐私需求增加，联邦学习在近年来受到关注，成为一大热点，，非常重要；
%  2. 联邦学习是一种新兴的分布式机器学习方法，可以保证在数据不离开设备或数据中心的情况下，让多个设备或数据中心协同地训练一个机器模型。联邦学习充分利用了各个设备或数据中心的数据，提高了模型的准确性和泛化能力，同时还保护了用户数据的隐私。

% ->like general FL（such as\ref{}） GFL:CS
%  3. 目前，联邦学习（介绍梯度聚合，衔接下文）通常采用的是 cs 架构，介绍CS架构（这样的架构好处是中央服务器可以集中控制模型训练的整个过程，并且可以使用高效的并行计算方法来提高计算效率；）->one of critical part 梯度聚合，简单介绍下现有工作（一句话概括一个method），

% 4. BUT 但是存在着中央服务器单点故障、网络拓扑可扩展性差等问题。受到 xxx 领域的启发(自然的)，p2p的优点是 鲁棒性\ref，可扩展性
%  5. 因此，本文考虑在引入decentralized架构下的图联邦学习。（转折）在这种框架下的联邦学习没有中央服务器的集中控制，有 、设计有效的模型聚合算法和交互规则等 的挑战或难点（比如，这种框架下，交互规则，梯度融合规则设计起来比较麻烦，不像cs架构那样直接等等）。

% 6. 在这篇文章中，为了应对这些挑战或难点，我们提出了 DGFL（full） 方法。（idea？？？？）该方法的idea是 每轮交互中用户采用随机单采样的方法发送消息，并引入了置信度的概念，根据置信度进行模型聚合。
% 我们首先做了 p2p引入到了图联邦（完全对等的学习节点组成，各学各的GNNs，no server）；然后基于linear的梯度聚合算法,linear的w怎样确定->conf->数据的simmilarity(基于相似度确定置信度，根据conf确定w)；最后做了实验，验证了所提方法的有效性，和FedAvg, Fedprox, GCFL and GCFL+方法进行了比较。
% 7. In summary, the main contributions of this paper are as follows: （1）贡献1：在decentralized 架构框架下提出了一种梯度交互机制，在保证xx前提下极大地降低通信开销；（2）贡献2：引入了节点之间的置信度（稍微扩一下，简单介绍，怎么计算），引入了一种新的基于提出的置信度的模型聚合方法（怎么聚合）（3）实验：在标准数据集上进行的实验，实验效果outperform。

% 8. The rest of this paper, The rest of this paper is organized as follows. Section 2 presents related work. Section 3 describes the preliminaries of GNNs and the most classic federation aggregation method. Section 4 provides details and the implementation of the proposed approach. Section 5 is the experiments to demonstrate the effectiveness of the proposed method. At last Section 6 provides the conclusion of this paper.

Graph learning refers to a class of methods that use the graph structure for machine learning. Specifically, graph learning uses information such as topology and neighbor relationships of graph structures to perform tasks such as feature learning, classification, and clustering of nodes or edges. Graph learning has a wide range of applications in healthcare\cite{wang2020moronet,jiang2020hi,covert2019temporal}, social network analysis\cite{hamilton2017inductive,xu2018powerful}, and intelligent transportation\cite{lv2020temporal,qiu2020topological,diao2022novel}, where the need for data privacy increases as well as the problem of data silos arises. Federated learning(FL) has received attention as a major hot topic in recent years. FL is an emerging distributed machine learning approach that ensures that multiple devices or data centers can collaboratively train a machine model without the data leaving the device or data center. FL leverages data from individual devices or data centers to improve the accuracy and generalization of the model, while also protecting the privacy of user data.

Like the classical FL framework, graph-federated learning(GFL) is also based on the FL framework. Currently, FL usually adopts the Client-Server(CS) framework, in which clients compute model parameters locally and upload them to the server, and the server aggregates the model parameters uploaded by each client and distributes them to the clients. One of the most important parts is the model aggregation method, such as FedAvg\cite{mcmahan2017communication}(which is based on a weighted average) and Fedprox\cite{li2020federated}(which adds regularization terms to the loss function).

However, the CS framework faces problems such as a single point of failure of the central server and poor scalability of network topology\cite{maly2003comparison,coulouris2005distributed}. Inspired by the blockchain domain, the decentralized framework has better robustness and scalability\cite{pilkington2016blockchain,raval2016decentralized}. Therefore, in this paper, we consider introducing the decentralized framework into GFL. Nevertheless, FL under this framework without centralized control of the server has challenges in designing reasonable interaction mechanisms and effective client-side model aggregation methods.

%Some sentences are awkward. For example, In abstract: "In this paper, we introduce the decentralized framework to graph-federated learning. First, we introduce the decentralized framework to graph-federated learning." This sounds redundant. In page2: .... ensuring no inefficiency. (2) introduces the confidence .... Probably, 'it' is necessary after (2) or change '.' to comma. 
In this paper, to address these challenges, we propose the Decentralized Graph-Federated Learning(DGFL) approach. First, we introduce the decentralized framework to GFL, which consists of fully peer-to-peer learning nodes without a central server. Each node has its own local data, aiming to train GNNs\cite{scarselli2008graph} models that are more suitable for local data. Second, determine the confidence among nodes based on the similarity of data among nodes, subsequently, the gradient information is then aggregated by linear weighting based on confidence. Finally, the proposed method is compared with FedAvg\cite{mcmahan2017communication}, Fedprox\cite{li2020federated}, GCFL, and GCFL+\cite{xie2021federated} to verify the effectiveness(accuracy, convergence speed, and computational time) of the proposed method. In summary, the main contributions of this paper are as follows: (1) a gradient interaction mechanism is proposed in the framework of decentralized architecture, which greatly reduces the communication overhead while ensuring no inefficiency. (2) introduces the confidence between nodes which is based on a local model gradient sequence and a new model gradient aggregation method based on the confidence for linear weighted aggregation is proposed. (3) experiments conducted on standard graph datasets, and the result of experiments demonstrate the proposed method outperforms other methods.

The rest of this paper is organized as follows. Section 2 presents related works. Section 3 describes the preliminaries of GNNs and the classic federated aggregation method. Section 4 provides details and the implementation of the proposed approach. Section 5 lists the experiments to demonstrate the effectiveness of the proposed method. At last, section 6 summarizes the entire article.

\section{Related Works}

\subsection{Federated Learning}

Federated Learning(FL)\cite{mcmahan2017communication} is a distributed machine learning framework that effectively helps multiple nodes or data centers to train models by performing machine learning. The participants of FL mainly consist of a central server and nodes. The server aims to train an optimal model based on the aggregated data of all parties, and the model trained by each node or data center serves local data goals. Importantly, FL allows participants (e.g., smartphones, sensors, mobile devices, servers, etc.) to not share data during the information exchange process so that the user's raw data remains local to the node or data center throughout the model training process, so FL simultaneously solves the problem of data silos while meeting the requirements of user privacy, data security, and government regulations. FL also reduces the pressure on network bandwidth because the local models are trained on local computers and only the model parameters need to be transmitted. The advantage of the CS framework is that it is easy to manage and easy to implement, and the centralized architecture makes it easier to control, coordinate and monitor. However, its centralized architecture will lead to a single point of failure, and once the server crashes, the whole system will not work properly and has poor scalability.

FL\cite{mcmahan2017communication} has three major elements: data source, federated learning system, and users. Under the federated learning system, each data source performs data preprocessing, jointly establishes its learning model, and feeds back the output results to the user. There are two main frameworks in FL:
\begin{itemize}
    \item \textit{client-server(CS) framework.} The client-server framework is the most commonly used framework for federated learning and was the first to be proposed. In this framework, there is a central server, which is responsible for coordinating the computation and communication among the various clients. Specifically, clients download models from the server, then train them using local data and upload the updated models to the server. The server aggregates all the models uploaded by the clients and calculates the average model, which is then sent back to the clients for the next round of training.

    \item \textit{decentralized framework.} The decentralized framework eliminates the reliance on a central server, and individual clients can communicate and collaborate directly with each other. In this framework, each client has its own model and uses local data for training, and then sends the updated model to other clients for model aggregation(such as gradient averaging). During model aggregation, clients can verify and monitor each other to ensure security and correctness. The decentralized framework has no single point of failure, each client is independent, and the system is more stable and robust. Data privacy is better because data does not need to be uploaded to the central server. However, the decentralized framework is relatively complex and requires more management and coordination efforts because there is no central server for management and control.
\end{itemize}

% Our work is based on FL under a decentralized network framework.

\subsection{Graph Learning}
A graph is a mathematical structure used to represent entities and their relationships, consisting of vertices and edges. Nodes represent entities (such as people, places, or objects), and edges represent relationships between entities (such as friendship, distance, or similarity). Machine learning on graphs is referred to as ``graph learning", and methods used in this field convert graph features to feature vectors of the same dimensionality in the embedding space. Without projecting the graph into a lower dimensional space, a graph learning model or algorithm directly converts graph data into the output of a graph learning architecture. Most graph learning approaches are based on or generalized from deep learning techniques since these techniques can encode and represent graph data as vectors. The objective of graph learning is to extract the desirable features of the graph, and the output vectors are in a continuous space. As a result, downstream activities like node classification and link prediction can employ graph representations with ease without using an explicit embedding procedure. Many graph analysis problems, including link prediction, recommendation, and classification, may be solved quickly and effectively in the representation space thanks to graph learning approaches\cite{guo2020graduate,bengio2013representation}. Different facets of social life, including communication patterns, community structures, and information diffusion, are shown by graph network representations\cite{xia2014exploiting,zhang2020data}. The three kinds of graph learning tasks—vertex-based, edge-based, and subgraph-based—can be separated depending on the characteristics of vertices, edges, and subgraphs. For categorization, risk identification, clustering, and community finding, a graph's vertex relationships can be leveraged\cite{xia2014community}. We can do recommendation and knowledge inference by determining whether an edge exists between any two vertices in a network.

\section{Preliminaries}
\subsection{Graph Neural Networks (GNNs)}

The core idea behind GNN\cite{scarselli2008graph} is to use the information interplay between neighbor nodes to replace the function illustration of every node and to procedure the total graph. Usually, a GNN model consists of a multi-layer neural network, and every layer of the community is accountable for updating the node function representation, and weighting and summing the characteristic vectors of the nodes to acquire a new characteristic illustration of the node. The shape and function data in the graph are normally expressed as $G=(V, E, X)$, where $V, E, X$ denotes nodes, links, and node features. Next, an illustration of an L-layer GNN is given as
\begin{equation}
    \label{gnn-v}
    h_v^{(l+1)} = \sigma(h_v^l, agg(\{h_u^l; u\in N_v\})), \forall l \in [L]
\end{equation}
the place $N_v$ is the set of neighbours of node $v$, $\sigma(\cdot)$ is the activation function, and the $h_v^l$ is the illustration of the node $v$ at the $l^{th}$-layer. $agg(\cdot)$ is the combination characteristic with special GNNs models. The equation \ref{gnn-v} is the node-level representation, and the graph-level illustration can be pooled from the illustration of all nodes, as
\begin{equation}
    h_G = readout(\{h_v; v\in V\})
\end{equation}
where the $readout(\cdot)$ represents the different pooling methods, such as mean pooling, sum pooling, etc, which is to aggregate and embed the vectors of each node in the graph into a single vector and then perform classification tasks.

\subsection{FedAvg}
FedAvg\cite{mcmahan2017communication} is a model parameter aggregation algorithm widely used in FL, which can be used to aggregate local model parameters from multiple clients into global model parameters. The main idea is to combine the parameters of the local model through the weighted average and use the weighted average parameters as the parameters of the global model. For example, there are $n$ clients in total, and $k$ is a specific communication round. The server will sample the client to select a set of participants $\{P\}_k$. For every participant $P_i$ in $\{P\}_k$, they use local data $D_i$ to train the model locally, and then pass its updated parameters $w_i^{(k)}$ to the server. The server then aggregates these incoming parameters by
\begin{equation}
    w^{(k+1)} = \sum_{i=1}^n\frac{|D_{i}|}{|D_{all}|}w_i^{(k)}
\end{equation}
where $|D_i|$ is the local data size of $P_i$, and $|D_{all}|$ represents the total data size of all $P_i\in \{P\}_k$. After the server aggregation update is completed, the new parameters of the global model will be passed to all clients.

\section{Decentralized Graph Federated Learning}
\subsection{The design of DGFL}

Suppose there is a set of local clients based on a decentralized network framework under the set $C=\{c_1, c_2, ..., c_n\}$, and the graph formed by the connections between the default clients is a fully connected graph. Unlike the classical CS framework, without the control of a central server, each local client needs to execute the federated algorithm once in each round of communication as a receiver. In contrast to Cluster Federated Learning (CFL)\cite{caldarola2021cluster}, which also does not rely on the central server to perform clustering, each local client decides its confidence weight based on the local data information and the messages received. Then, clients aggregate the sender's gradient information received in this round with the local gradient information in this round according to the confidence weight and then update the local gradient information. 
\begin{figure}[htbp]
	\centering
	\includegraphics[scale=0.4]{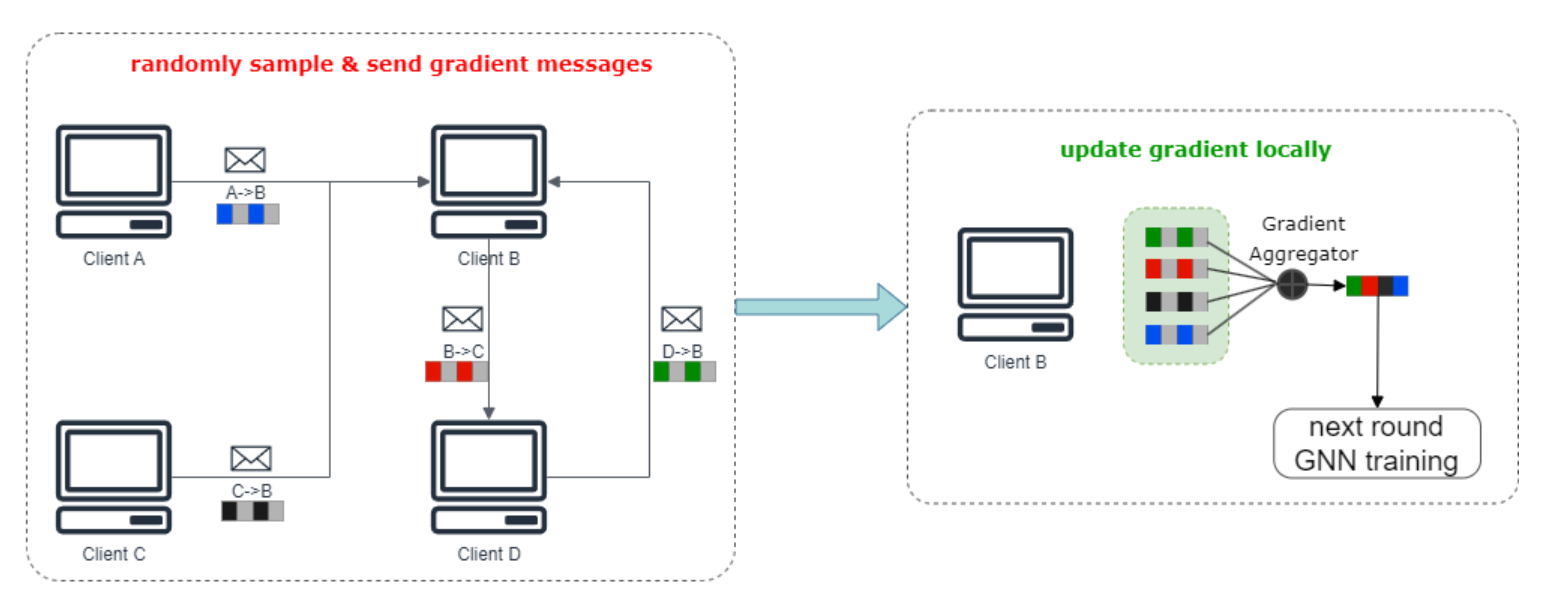}
	\caption{the design of the proposed method}
	\label{flow}
\end{figure}

Figure \ref{flow}, which is a random situation in the interaction process, describes the main steps of the proposed algorithm. The left part of the figure shows the process of randomly sampling and sending messages to the selected client. Take client B as an example, the right part of the figure demonstrates the gradient updating process.

The definition of the local client set has been given: 
$C=\{c_1, c_2, ..., c_n\}$, assuming that each client maintains a local neighbor set $N_i = \{n_1, n_2, n_3, ..., n_k\}$, where $N_i$ represents the neighbor list of $c_i$. From the above process, we can know that in each round of communication, when a client acts as the sender, it will use random single sampling which randomly selects a neighbor as the message receiver of the current round. Of course, each client also needs to maintain a set of senders $S_i=\{s_1, s_2, s_3, ..., s_k\}$, where $S_i$ represents the sender list of $C_i$, which is used to record who send a message to yourself in this round. Then the client $C_i$ selects the senders greater than the average confidence value from the sender list to aggregate their messages. Next, define confidence($Conf$) as
\begin{equation}
	\label{conf}
	Conf_{i,j}^t = 1-DTW_{std}(g_i^t,g_j^t) 
\end{equation}

The average confidence value($\overline{Conf^t}$) in the t-th round is denoted as
\begin{equation}
    \label{avgconf}
    \overline{Conf_i^t}=\frac{1}{|S_i|}\sum_{k\in S_i}Conf_{i,k}^t
\end{equation}
$Conf_{i,j}^t$ represents the confidence of client $c_i$ on $c_j$ during the t-th round of communication, and $Conf\in [0, 1]$, where $DTW_{std}$ represents the standardized $DTW()$ function which is used to calculate the dynamic time wrapping (DTW)\cite{muller2007dynamic} distance between two gradient sequences. Obviously, we can get $Conf_{i, i} = 1$.

It should be emphasized that $DGFL$ is different from the classic federated algorithm. What the client updates after each round of communication is no longer the model parameters, but the gradient. The specific equation of the strategy is as follows:
\begin{equation}
	g_i^{(t+1)} = \frac{Conf^t_{i,i}*g_i^t+\sum_{j\in Sample}Conf^t_{i,j}*g_{j}^t}{Conf^t_{i,i}+\sum_{j\in Sample}Conf^t_{i,j}}
	\label{gra_aggr}
\end{equation}
The Equation \ref{gra_aggr} is the gradient update strategy of the client $c_i$, where $t$ represents the communication round, $g$ is the gradient, and $Sample$ is the selected senders after a de-mean confidence filtering. At the same time, the local and received gradient information is normalized. 

\subsection{Algorithm implementation}
The implementation of DGFL is illustrated in algorithm \ref{alg}. 
Lines 1 to 5 show the preparations before starting which initialize all clients. From line 6 to the end is the FL process. In the process of FL, sending process is from lines 7 to 12, and the receiving process is from lines 13 to 20. The random single-sampling process is shown in line 9. The computation of confidence between clients on line 15 is based on equation \ref{conf}. The gradient updating is shown on line 20.

\begin{algorithm}[H]
\caption{Decentralized Graph Federated Learning}
\label{alg}
\SetKwInOut{Input}{Initialization}\SetKwFunction{LocalTrain}{ClientLocalTrain}\SetKwData{Send}{send}\SetKwData{Select}{select}\SetKwData{Compute}{compute}\SetKwData{Append}{append}\SetKwData{Update}{update}

\Input{Set clients $C=\{c_1, c_2, ..., c_n\}$, divide the dataset unevenly among clients, $E$ is the number of local epochs, $n$ is the total number of clients}
% \tcp{Preparations before starting}
\For{each client $k \in C$}{
$S_k \leftarrow \{\}$\;
$N_k \leftarrow C\backslash c_k$\;
$Sample \leftarrow \{\}$
}
% \tcp{Start federated learning}
\For{each round t = 1, 2 ,3, ...}{
    % \tcp{client acts as a sender}
    \For{each client $i \in C$}{
        $g_i^t \leftarrow$ \LocalTrain{E}\;
        $r \leftarrow $randomly \Select a client from $N_i$\;
        $S_r \leftarrow i$(\Append client $i$ \KwTo client $r$'s sender list)\;
        \Send $g_k^t$ \KwTo r
    }
    % \tcp{client acts as a recipient}
    \For{each client $i \in C$}{
        \For{$j \in S_i$}{
            \Compute $Conf_{i,j}^t$
        }
        
        $Sample\leftarrow j\in S_i$, for $Conf_{i,j}^t\geq \overline{Conf_i^t}$(\Append client $j$ \KwTo sample list)\;
        $g_i^{(t+1)} \leftarrow \frac{Conf^t_{i,i}*g_i^t+\sum_{j\in Sample}Conf^t_{i,j}*g_{j}^t}{Conf^t_{i,i}+\sum_{j\in Sample}Conf^t_{i,j}}$(\Update gradient)\;
        $Sample\leftarrow \{\}$
    }
}
\end{algorithm}

\section{Experiments}

\subsection{Experimental settings}
\paragraph{Datasets} The datasets we use is TuDataset from the official datasets set of PyG (PyTorch Geometric), which mainly selects different data sets from three fields, namely Small molecules (AIDS, DHFR, P388), Bioinformatics (DD), Social networks(COLLAB, IMDB-BINARY, IMDB-MULTI), where each dataset has a set of graphs. The graph labels are binary or multi-class and our task is a graph classification task. We randomly distribute the graphs of a single dataset to multiple clients and keep 10\% of the graphs as the test set.
\begin{table}[htbp]
\centering
\caption{Data Set Stat.}
\label{conv}
\begin{tabular}{|l|l|l|l|l|}
\hline
Name & \textit{Graphs} & \textit{Classes} & \textit{Avg.nodes} & \textit{Avg.edges}  \\
\hline
COLLAB & 5000 & 3 &74.49 & 2457.78 \\
AIDS & 	2000 & 2 & 15.69 & 16.20 \\
DD & 1178 & 2 & 284.32 & 715.66 \\
PTC\_FR & 351 & 2 & 14.56 & 15.00 \\
IMDB-BINARY & 1000 & 2& 19.77 & 96.53\\
\hline
\end{tabular}
\end{table}

\paragraph{Parameters} The local epoch $E$ is set to 5 for each different FL algorithm. In the neural network layer, we choose the 3-layer Graph Isomorphism Networks(GINs) with a hidden size of 64. The batch size is set to 128 and the optimizer is Adam\cite{kingma2014adam} with weight decay of $5e^{-4}$ and a learning rate of 0.001. The parameter $\mu$ of  Fedprox\cite{li2020federated} is set to 0.01. All experiments are run on the server with  16GB NVIDIA Tesla T4 GPUs.

\paragraph{Topology} Like the decentralized framework introduced by DGFL in Chapter 4, each node in the experiment is composed of a fully connected graph, meaning that each node has communication channels with all other nodes. In the experiment, we set the number of clients to 10 by default.

\subsection{Research questions}
In this section. we will mainly conduct experiments to address the following research questions:
\begin{itemize}
    \item How do the convergence speed, loss, and accuracy of the proposed method compare with other methods (Effectiveness) 
    
    \item How does the time cost of the proposed method compare with other methods? (Time Overhead) 
\end{itemize}
\textbf{Effectiveness.} To illustrate the superiority of the proposed method in terms of effectiveness, we mainly compare the proposed method with the existing baseline methods from three aspects which are accuracy, loss, and convergence speed.

\begin{figure}[htbp]
	\centering
	\includegraphics[width=0.5\textwidth]{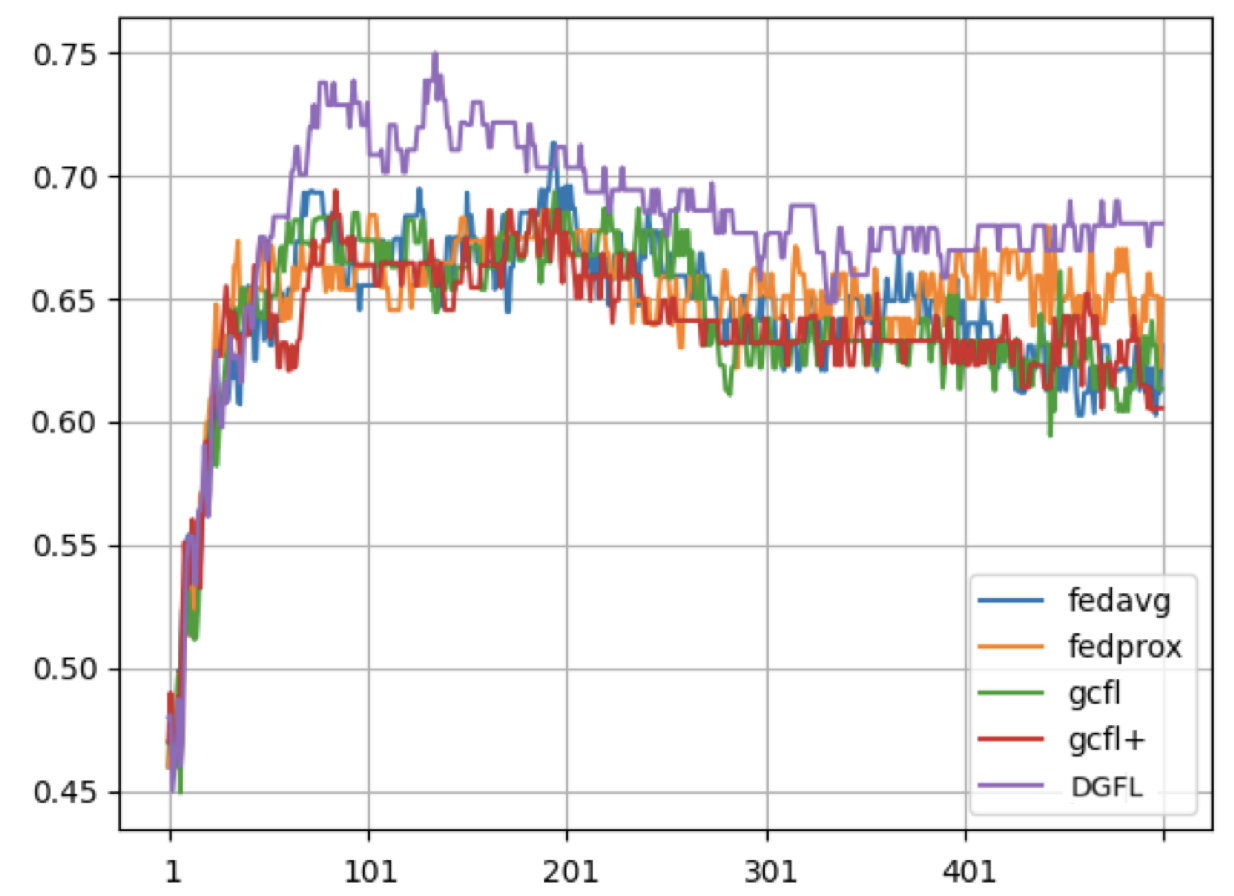}
	\caption{500 rounds accuracy comparison on IMDB-BINARY}
	\label{binary}
\end{figure}

\paragraph{Accuracy.} Figure \ref{binary} shows the accuracy variation of the proposed method compared to the baseline method over 500 rounds for the same dataset IMDB-BINARY. Table \ref{acc} shows the final accuracy comparison of all methods with different data sets for a fixed number of rounds. The horizontal axis is the exchange rounds, and the vertical axis represents the average accuracy of all clients.

\begin{table}[htbp]
\centering
\caption{Accuracy comparison of all methods for a fixed number of rounds with different datasets.}\label{acc}
\begin{tabular}{|l|l|l|l|l|l|}
\hline
    & COLLAB$_{500}$ & AIDS$_{500}$ & DD$_{1000}$ & PTC\_FR$_{1000}$ & IMDB-B$_{1000}$ \\
\hline
   FedAvg\cite{mcmahan2017communication}  & 0.7253 & 0.9859 & 0.7277 & 0.7250 & 0.6329\\
\hline
   Fedprox\cite{li2020federated}  & 0.7141 & 0.9858 & 0.7039 & 0.7250 & 0.6540\\
\hline
   GCFL\cite{xie2021federated}  & 0.6879 & 0.9808  & 0.7369 & 0.7250 & 0.6613\\
\hline
   GCFL+\cite{xie2021federated}  & 0.6962 & 0.9808 & 0.7181 & 0.7250 & 0.6511\\
\hline
   DGFL  & \textbf{0.7374} & \textbf{0.9906}  & \textbf{0.7628} & \textbf{0.7500} & \textbf{0.6732}\\
\hline
\end{tabular}
\end{table}

\begin{figure}[htbp]
\centering
\subfigure[AIDS(200)]{
\includegraphics[width=0.28\textwidth]{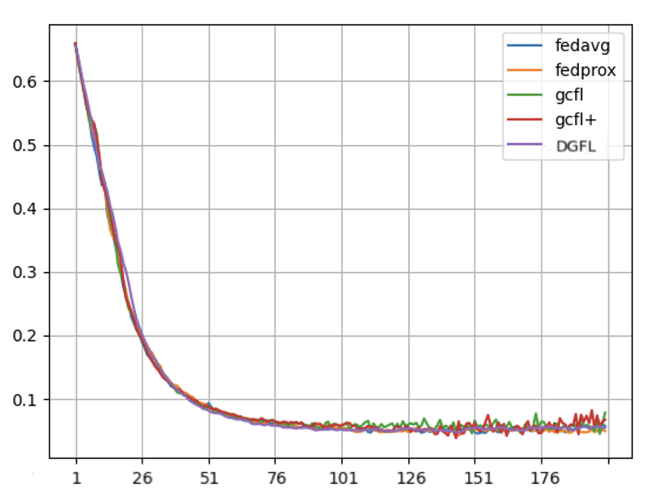} \label{loss1}
}
\subfigure[DHFR(100)]{
\includegraphics[width=0.29\textwidth]{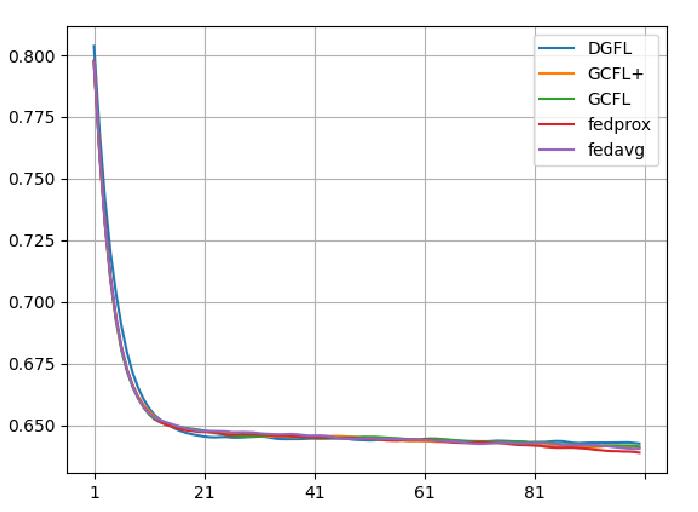} \label{loss2} 
}
\subfigure[AIDS(1000)]{
\includegraphics[width=0.308\textwidth]{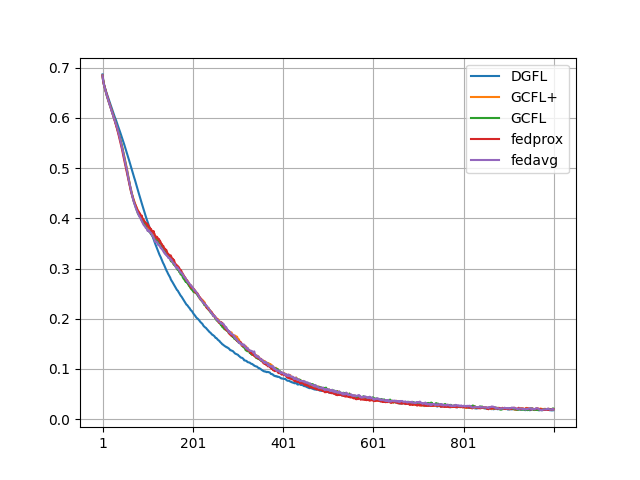} \label{loss3} 
}

\caption{Loss comparison of methods.}
\end{figure}

\paragraph{Loss.} Here are the results of three comparative experiments of different methods based on the loss to demonstrate the advantages of the proposed method. The vertical axis of each graph is the loss, and the horizontal axis is the communication round. From the first result \ref{loss1} of the experiment on dataset AIDS of 200 rounds, we can find that the loss performance of all methods is very similar, however, compared with the baseline methods, the proposed method has better stability, which means that the baseline method has a larger "shake range". After that, the number of AC rounds is increased to 1000, and it can be seen from \ref{loss3} that the proposed method has the highest curvature of the convergence curve in the case where the final convergence of the losses of each method is almost the same.

\begin{table}[htbp]
\centering
\caption{Convergence comparison of methods on AIDS.}\label{conv}
\begin{tabular}{|l|l|l|l|l|l|}
\hline
Method Name& FedAvg\cite{mcmahan2017communication} & Fedprox\cite{li2020federated} & GCFL\cite{xie2021federated} & GCFL+\cite{xie2021federated} & DGFL \\
\hline
Convergence round & 65 & 62 & ``63" & ``61" & \textbf{47}\\
\hline
\end{tabular}
\end{table}

\begin{figure}[htbp]
	\centering
	\includegraphics[width=0.5\textwidth]{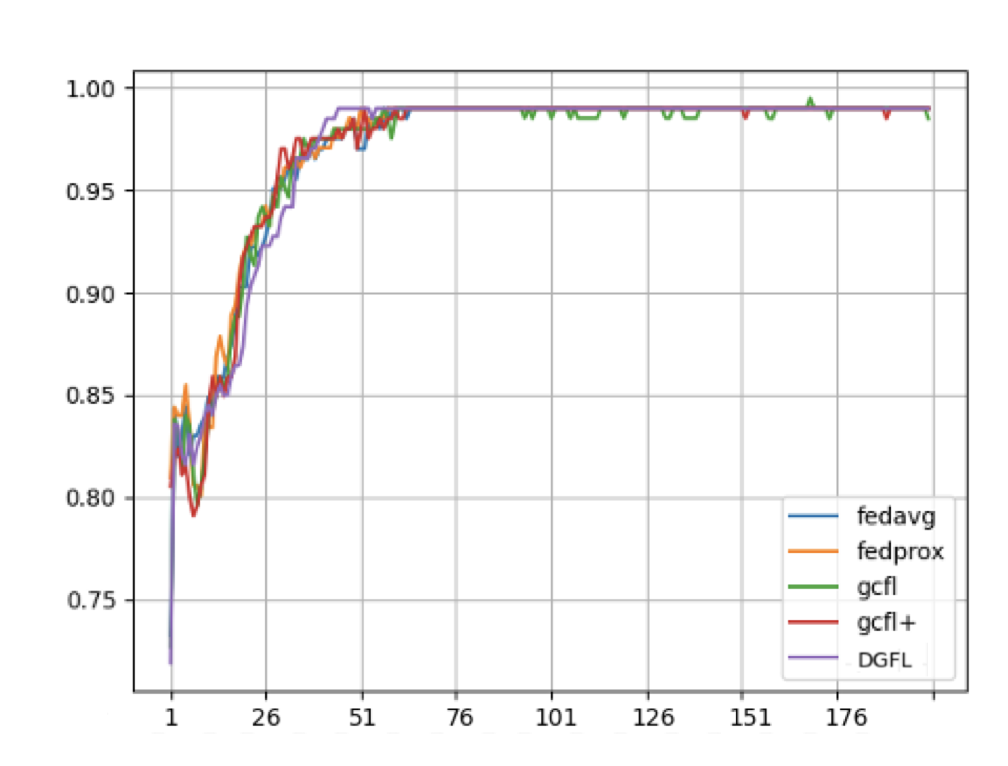}
	\caption{200 rounds convergence speed comparison on AIDS.}
	\label{aids}
\end{figure}

\paragraph{Convergence speed.} When the fluctuation range of the accuracy within a certain number of rounds remains within a certain threshold, we consider that the method has reached convergence. Since most datasets cannot achieve convergence within a short communication round, we give the dataset AIDS where all methods can converge within a short communication round for comparative experiments, and the results are shown in Figure \ref{aids}.

Obviously, in Table \ref{conv}, we can see that the number of convergence rounds of other baseline methods is around 60, while the proposed method is reduced below 50 which is a big boost. The reason why we put quotation marks on the convergence round of $GCFL$ and $GCFL+$ for these two methods is that they have not reached absolute convergence. It can be seen that there is still a slight shake in the follow-up.\\

\textbf{Time overhead.} Under the premise of ensuring that the proposed method has an advantage in accuracy, compare the time consumption with the existing baseline method under the same dataset and the same number of rounds.
\begin{table}[htbp]
\centering
\caption{Time consumption comparison of methods for a fixed number of rounds with different datasets.}\label{time}
\begin{tabular}{|l|l|l|l|l|l|}
\hline
   \quad  & $COLLAB_{500}$ & $AIDS_{500}$ & $DD_{1000}$  & $IMDB-B_{1000}$ \\
     
\hline
   FedAvg\cite{mcmahan2017communication}  & 6467s & 842s & 2552s  & 1475s\\
\hline
   Fedprox\cite{li2020federated}  & 6153s & 998s & 2584s & 1591s\\
\hline
   GCFL\cite{xie2021federated}  & \textbf{5843s} & 887s  & 2751s & 1373s\\
\hline
   GCFL+\cite{xie2021federated}  & 6374s & 867s & 2568s  & 1279s\\
\hline
   DGFL  & 6012s & \textbf{838s}  & \textbf{2270s}  & \textbf{1128s}\\
\hline
\end{tabular}
\end{table}

It can be seen from Table \ref{time} that except for the time consumption of the proposed method under COLLAB(500 rounds) is slightly higher than that of GCFL, the time consumption of the proposed method under other datasets is optimal. Moreover, with the number of communication rounds increasing, the proposed method has no disadvantage in accuracy, and the speed of training highlights a greater advantage.

\section{Conclusion}

% 两句话讲本文解决的问题，核心思路，提出的方法
Taking the interaction of graph-federated learning based on a decentralized structure into consideration, this paper adopted GIN and a new federated aggregator algorithm, by introducing the proposed algorithm. 
% 一句话讲实验场景
Following, we considered four algorithms (i.e., FedAvg, Fedprox, GCFL, and GCFL+), and conducted a set of experiments to validate the effectiveness of the proposed algorithms. 
%  说我们的方法很强，有前景。
The experimental results illustrated that the proposed algorithms could achieve graph-federated learning\cite{mcmahan2017communication} based on a decentralized structure.
The success of graph learning and federated learning in this area demonstrates its great potential for other applications.

% 简单总结两条不足。
However, heterogeneity of each client's local data structure (e.g. distributing each client's different local data from different datasets) and Byzantine problems in communication (e.g. a certain client intentionally sends false information to other clients) will affect the training results. 
Therefore, there is still a need to address the problem of working under more dynamic topological conditions.
% 此外还要做的future work是啥
Furthermore, the success of the proposed algorithm in graph-federated learning based on a decentralized framework shows the potential to be applied to more real-world scenarios, such as medical institutions sharing patient data for model training. In the future, we aim to improve the deficiencies in the proposed algorithm and apply it to more realistic scenarios.

\bibliographystyle{splncs04}
\bibliography{ref}

\end{document}